# The Simplest Balance Controller for Dynamic Walking

Linqi Ye, Xueqian Wang, Houde Liu, Bin Liang

*Abstract*—Humans can balance very well during walking, even when perturbed. But it seems difficult to achieve robust walking for bipedal robots. Here we describe the simplest balance controller that leads to robust walking for a linear inverted pendulum (LIP) model. The main idea is to use a linear function of the body velocity to determine the next foot placement, which we call linear foot placement control (LFPC). By using the Poincaré map, a balance criterion is derived, which shows that LFPC is stable when the velocity-feedback coefficient is located in a certain range. And that range is much bigger when stepping faster, which indicates "faster stepping, easier to balance". We show that various gaits can be generated by adjusting the controller parameters in LFPC. Particularly, a dead-beat controller is discovered that can lead to steady-state walking in just one step. The effectiveness of LFPC is verified through Matlab simulation as well as V-REP simulation for both 2D and 3D walking. The main feature of LFPC is its simplicity and inherent robustness, which may help us understand the essence of how to maintain balance in dynamic walking.

## I. INTRODUCTION

Robust balance is one of the most important goals in the study of bipedal robots. Perhaps we can learn how to achieve that goal from human beings since human can recover balance very well when perturbed. Generally, there are three ways that human takes to recover balance [1]: 1) Taking a step to move the zero moment point within the base of support; 2) Using ankle torques to move the center of pressure of the feet; 3) Distorting the upper body to change angular momentum. Among them, the last two strategies seem less efficient and can only reject small disturbances. We can imagine the case of standing still and being pushed. When pushing slightly, we can recover balance using strategy 2 or 3 without stepping. But for a strong push, the only way to avoid a fall is to take one or more steps in the pushing direction. This indicates that stepping is central for achieving robust balance. Therefore, how to select the foot placement then becomes the key issue. Intuitively, it is natural to think that we should take steps in the direction of falling to prevent a fall. However, it seems not clear yet how we exactly selects the step location. According to our experience, the step length we take is possibly related to the falling speed, that is, when falling faster, we tend to take bigger steps.

In the robotics field, foot placement control also has been widely studied. Some high-dynamic robots have had balance control based almost entirely on foot placement. The best known is from Marc Raibert's MIT lab and his company Boston Dynamics. During Raibert's early research of a hopping robot [2], an interesting observation was found, that is, when the robot places its foot on a special location called "neutral point", it will move on a symmetric trajectory and keeps its forward velocity unchanged. Moreover, the robot speeds up when placing its foot ahead of the neutral point and slows down when placing its foot behind the neutral point. Using this property, Raibert designed a simple foot placement controller [2] which is known as "Raibert heuristic" and has been applied to various legged systems range from monopod, biped to quadruped robots [2-4]. The stability analysis and attraction of domain for "Raibert heuristic" can be found in [5]. Another approach to control using foot placement is the capture point [6-8] method developed by Jerry Pratt. Capture point refers to the point on the ground where a robot must step to in order to come to a complete stop. For a LIP model, the position of the capture point is determined by the body velocity multiplied by a constant coefficient. From "Raibert heuristic" to capture point, we find a common feature, that is, both of them have applied linear functions of the body velocity for foot placement. Besides, another famous biped control method named "SIMBICON" [9] has also adopted a similar balance strategy together with a finite state machine to generate a large variety of gaits, including walking, running, skipping, and hopping. And the authors have demonstrated that those gaits are very robust to disturbances through physics-based simulations. Recently, a discrete PD controller for foot placement was proposed in [10] and applied to the ATRIAS robot. The stability has been analyzed based on the spring-loaded inverted pendulum (SLIP) model.

All of the aforementioned work have shown the feasibility of LFPC from both theoretical analysis and practical applications. In this paper, we do theoretic analysis on LFPC using the LIP model [11], which has not yet been done in the existing literature as far as we know. Due to the simplicity of the LIP model, the stability condition and periodic solutions can be explicitly derived, which is very helpful for us to further understand LFPC. We also found that the capture point can be viewed as a particular case of LFPC and discovered a dead-beat controller that can lead to faster convergence than the capture-point controller.

The contribution of this paper is that we present and comprehensively analyze the simplest balance controller based on LFPC for both 2D and 3D LIP models. There are numerous papers published on walking dynamics and balance

This work was supported by National Natural Science Foundation of China (62003188, U1813216), Shenzhen Science Fund for Distinguished Young Scholars (RCJC20210706091946001), and Guangdong Special Branch Plan for Young Talent with Scientific and Technological Innovation (2019TQ05Z111).

L. Ye is with (1) Artificial Intelligence Institute, Shanghai University, Shanghai 200444, China; (2) Center of Intelligent Control and Telescience, Tsinghua Shenzhen International Graduate School, Tsinghua University, Shenzhen 518055, China (e-mail: ye.linqi@sz.tsinghua.edu.cn).

X. Wang, and H. Liu are with the Center of Intelligent Control and Telescience, Tsinghua Shenzhen International Graduate School, Tsinghua University, 518055 Shenzhen, China (e-mail: { wang.xq, liu.hd}@sz.tsinghua.edu.cn).

B. Liang is with the Navigation and Control Research Center, Department of Automation, Tsinghua University, 100084 Beijing, China (e-mail: bliang@tsinghua.edu.cn).

control, such as the work on capture steps [12], divergent component of motion [13], model predictive control [14], self-stabilization walking [15], robust control [16], and so on. However, it is difficult to obtain an intuitive understanding on how to maintain balance during walking from existing work. Therefore, the results in this paper can be a useful complement to the field of legged locomotion.

The rest of this paper is organized as follows. In Section II, we introduce the LFPC strategy in 2D and perform theoretical analysis. In section III, LFPC is extended to 3D. Then simulation results are given in Section IV and conclusions are given in Section V.

## II. LINEAR FOOT PLACEMENT CONTROL IN 2D

### A. LFPC Description

The LIP walking model is depicted in Fig. 1. The model has a point mass on the hip and two massless legs with point feet. During walking, the center of mass (CoM) maintains a constant height. Leg transition happens in a smooth way where the body velocity maintains the same. We assume an instantaneous double stance and no slipping of the feet.

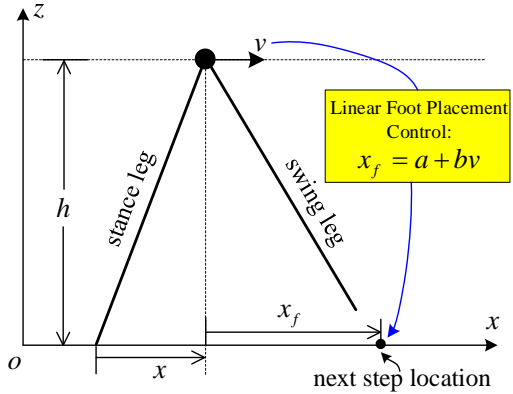

Figure 1. Linear foot placement control for LIP walking. The next step location is determined by a linear function of the body velocity.

During the continuous phase of a step, the system follows the dynamics of a linear inverted pendulum, which is [11]

$$\ddot{x} = gx/h \quad (1)$$

where $x$ is the horizontal position of the CoM relative to the stance foot, $g$ is the acceleration of gravity, and $h$ is the body height.

The solution of (1) can be calculated as follows (an easy way to do this is to use symbolic computations by Matlab, and in the subsequent calculations, we have highly relied on that)

$$x(t) = x(0)\cosh(t/T_c) + T_c \dot{x}(0)\sinh(t/T_c)$$
$$\dot{x}(t) = x(0)\sinh(t/T_c)/T_c + \dot{x}(0)\cosh(t/T_c) \quad (2)$$

where $T_c = \sqrt{h/g}$.

As shown in Fig. 1, the main idea of LFPC is to select the next step location as a linear function of the body velocity, that is

$$x_f = a + bv \quad (3)$$

where $x_f$ is the position of the next step location relative to the CoM, $v$ is the body velocity at the moment of touch down, and $a, b$ are controller parameters.

Equation (3) tells us where to take the next step, which uses only two parameters. Meanwhile, we also need to specify when to take the next step. Here we simply use a time duration $T$ to trigger the swing foot touch down.

### B. The Balance Criterion

Suppose the initial state is $(x_0, v_0)$ in the beginning of the first step. According to (2), the state just before swing foot touch down can be calculated as

$$x_1^- = x_0 c_T + T_c v_0 s_T$$
$$v_1^- = x_0 s_T / T_c + v_0 c_T \quad (4)$$

where

$$s_T = \sinh(T/T_c), c_T = \cosh(T/T_c) \quad (5)$$

with $T$ being the step period.

When touch down finishes, the swing leg becomes the new stance leg and the next step begins. Denote the initial state of the next step as $(x_1, v_1)$. Since the body velocity keeps unchanged during transition and $x_1$ is the body position relative to the stance foot, we have

$$v_1 = v_1^-, \ x_1 = -x_f \quad (6)$$

Combining (3), (4), and (6), it follows that

$$x_1 = -a - b(x_0 s_T / T_c + v_0 c_T)$$
$$v_1 = x_0 s_T / T_c + v_0 c_T \quad (7)$$

which is the Poincaré map [17] of the system.

For the Poincaré map (7), denote it as $f(q)$, where $q = [x_0, v_0]^T$. Then the Jacobian of $f(q)$, i.e., $\partial f / \partial q$, can be obtained as follows

$$J = \begin{bmatrix} -bs_T/T_c & -bc_T \\ s_T/T_c & c_T \end{bmatrix} \quad (8)$$

The eigenvalues of $J$ is

$$\lambda_1 = 0, \ \lambda_2 = c_T - bs_T/T_c \quad (9)$$

To maintain stability, it requires that all the eigenvalues stay in a unit circle, that is

$$|c_T - bs_T/T_c| < 1 \quad (10)$$

which can be rewritten as

$$T_c \frac{c_T - 1}{s_T} < b < T_c \frac{c_T + 1}{s_T} \quad (11)$$

Equation (11) is the stability condition of LFPC. Therefore, we call it the *balance criterion*.

A direct interpretation of (11) is that the velocity feedback coefficient should be located in a certain range which is determined by the system parameter $T_c$ and the step period $T$. It also indicates that the stability has nothing to do with the parameter $a$. Another important observation is that the Jacobian (8) is independent of the system state, which indicates that the system is globally stable once the stability condition (11) is satisfied. This is an important feature of LFPC, which reflects its inherent robustness.

Equation (11) can also be written as

$$\frac{c_T - 1}{s_T} < \frac{b}{T_c} < \frac{c_T + 1}{s_T} \quad (12)$$

With (12), the stability condition is represented by the colored area in Fig. 2.

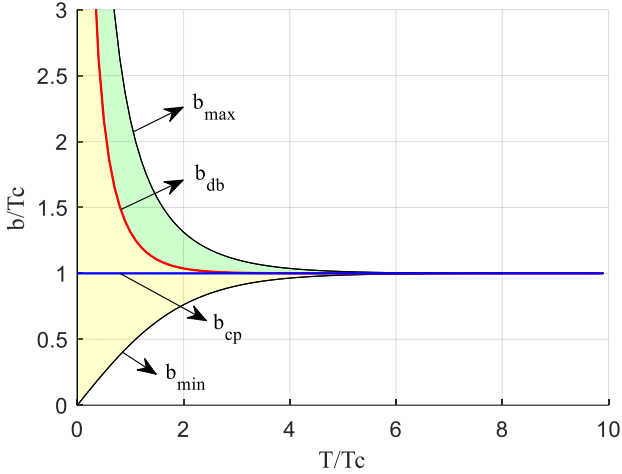

Figure 2. Balance criterion of LFPC. The colored area represents the parameter space that leads to stable walking. The red line is the "dead-beat" controller that converges in one step. The green area is the underdamped area (converge with oscillation) and the yellow area is the overdamped area (no oscillation). The capture-point parameter $b = T_c$ (the blue line) is a special case in the overdamped area.

From Fig. 2, we find the balance criterion can be summarized as "smaller $T$, wider range of $b$". Particularly, there is a constant parameter $b = T_c$ that maintains stability for all values of $T$. This is exactly the feedback parameter that corresponds to the capture point [6].

For convenience of further analysis, let $a = 0$ and assume the initial value satisfies $x_0 = -x_f = -bv_0$, then we have

$$v_1 = (-b s_T / T_c + c_T) v_0 \quad (13)$$

Based on this, the stability region in Fig. 2 is further analyzed as follows.

(1) The upper bound $b_{max} = T_c(c_T + 1)/s_T$

In this case, we have $\lambda_2 = -1$ and $v_1 = -v_0$. If the robot steps to this point, it will recover to its initial position with reversed velocity. This is the farthest point that the robot can step to where its velocity will not increase in the end of the step.

(2) The lower bound $b_{min} = T_c(c_T - 1)/s_T$

In this case, we have $\lambda_2 = 1$ and $v_1 = v_0$. If the robot steps to this point, it will go over this point and reach the mirror position with the same velocity. This is the closest point that the robot can step to where its velocity will not increase in the end of the step.

(3) The dead-beat parameter $b_{db} = T_c c_T / s_T$

In this case, we have $\lambda_2 = 0$ and $v_1 = 0$. If the robot steps to this point, it will stop above this point in the end of this step. This is the point that the robot can immediately stop (in one step), which we call the *dead-beat point* (see [18] for meaning of "dead-beat").

(4) The capture-point parameter $b_{cp} = T_c$

This is an ordinary as well as a special value in the overdamped area. It is ordinary since the robot stops gradually just like the other points in the overdamped area. It is special since this is the only constant value independent of $T$ that leads to stable walking. In addition, the robot will stop over this point when $T$ is infinity, which is the reason that [6] defines it as the capture point.

For the readers' convenience, Fig. 3 has given an example showing the trajectories of the CoM position and velocity.

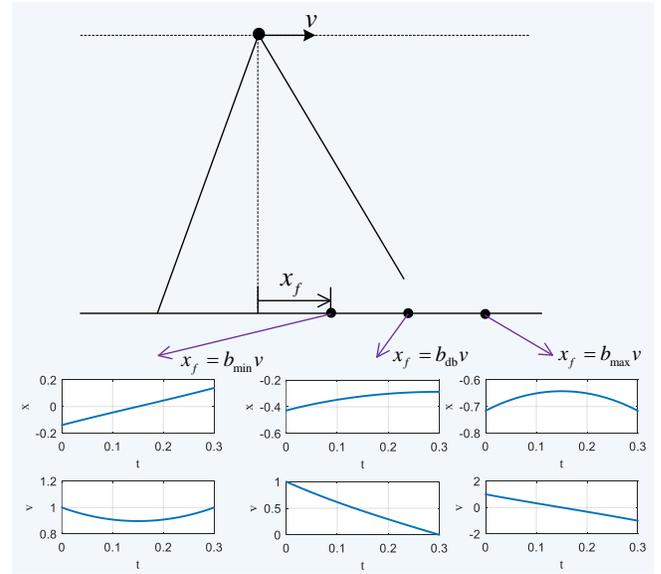

Figure 3. Example trajectories of the CoM position and velocity when stepping to different location. The time starts at touch down and ends at 0.3 s.

The colored area in Fig. 2 ($b_{min} < b < b_{max}$) is the converging region, where the robot will decelerate. Out of this area, the robot will accelerate in the end of the step. When a robot is perturbed, it is likely to fall due to a large velocity. Therefore, the range of ($b_{min}, b_{max}$) reflects the feasible region that the robot can step to to maintain balance, which can be used as a metric of robustness. Since the range ($b_{min}, b_{max}$) is much bigger for a smaller step period, it indicates that the robot is more robust when stepping faster. We summarize this property as "**Faster stepping, easier to balance**".

In the overdamped area, where $b_{db} < b < b_{max}$ ($\lambda_2 \in (0,1)$), the robot will decelerate with the velocity direction unchanged at each step and stop gradually. In the underdamped area, where $b_{min} < b < b_{db}$ ($\lambda_2 \in (-1, 0)$), the robot will decelerate with the velocity direction reversed at each step and stop gradually.

**Comment 1**: The balance criterion (11) is derived for constant values of the parameters $T, a, b$. It is a sufficient but not necessary condition to maintain stability when the parameters are varying in each step. For example, the robot can accelerate first and then decelerate before it stops.

**Comment 2**: For $a \neq 0$, the steady-state velocity is not zero. In this case, the properties above still hold for the velocity deviation $e = v - v^*$ ($v^*$ is the steady-state velocity).

*C. Periodic Solutions*

After obtaining the balance criterion, another question is where the state will converge to during stable walking. Answering this question relies on finding the fixed points of the Poincaré map.

**(1) Period-1 gait**

When both legs use the same control parameters $a, b$, the period-1 gait emerges. In a period-1 gait, the initial state of each step is the same, so we have

$$x_1 = x_0, v_1 = v_0 \quad (14)$$

Combing (7) and (14) leads to the following equation set

$$-a - b(x_0 s_T / T_c + v_0 c_T) = x_0$$
$$x_0 s_T / T_c + v_0 c_T = v_0 \quad (15)$$

By solving (15) we obtain

$$x_0 = \frac{aT_c(c_T - 1)}{T_c - T_c c_T + b s_T}$$
$$v_0 = \frac{-a s_T}{T_c - T_c c_T + b s_T} \quad (16)$$

Equation (16) is the solution of the period-1 gait. The step length (defined as the next step position minus the stance foot position, it can be negative which indicates walking in the $-x$ direction) can be obtained as follows

$$d = -2x_0 = \frac{2aT_c(1 - c_T)}{T_c - T_c c_T + b s_T} \quad (17)$$

It can be seen from (17) that the step length is proportional to $a$. For a fixed $b$ and $T$, bigger $a$ indicates a bigger step length as well as a higher walking speed (the average walking speed is $d / T$). Particularly, when $a = 0$, the robot will stop and step in-place.

**(2) Period-2 Gait**

When the two legs use different control parameters, we can obtain a period-2 gait, that is, the initial state recovers to $(x_0, v_0)$ after taking two steps.

Denote the controller parameters of leg 1 and leg 2 as $a_1, b_1$ and $a_2, b_2$, respectively (it means that when leg 1 swings, it uses the parameter $a_1, b_1$ and when leg 2 swings, it uses the parameter $a_2, b_2$). Assume leg 1 swings first and then leg 2 swings. Denote the initial state of the first step (leg 1 swings) as $(x_0, v_0)$, the initial state of the second step (leg 2 swings) as $(x_1, v_1)$, and the initial state of the third step (leg 1 swings) as $(x_2, v_2)$. Following (7), we can easily obtain the following relationship

$$x_1 = -a_1 - b_1(x_0 s_T / T_c + v_0 c_T)$$
$$v_1 = x_0 s_T / T_c + v_0 c_T \quad (18)$$

$$x_2 = -a_2 - b_2(x_1 s_T / T_c + v_1 c_T)$$
$$v_2 = x_1 s_T / T_c + v_1 c_T \quad (19)$$

Substituting (18) into (19) and combining $x_2 = x_0, v_2 = v_0$ it follows that

$$x_0 = -a_2 - b_2 \left[ -a_1 - b_1(x_0 s_T / T_c + v_0 c_T) \right] s_T / T_c$$
$$\quad - b_2 c_T (x_0 s_T / T_c + v_0 c_T)$$
$$v_0 = \left[ -a_1 - b_1(x_0 s_T / T_c + v_0 c_T) \right] s_T / T_c +$$
$$\quad c_T (x_0 s_T / T_c + v_0 c_T) \quad (20)$$

The solution of (20) is

$$x_0 = \frac{-T_c(T_c a_2 - a_1 b_2 s_T - T_c a_2 c_T^2 + a_2 b_1 c_T s_T)}{T_c^2 - T_c^2 c_T^2 + T_c b_1 c_T s_T + T_c b_2 c_T s_T - b_1 b_2 s_T^2}$$
$$v_0 = -\frac{T_c a_1 s_T + T_c a_2 c_T s_T - a_2 b_1 s_T^2}{T_c^2 - T_c^2 c_T^2 + T_c b_1 c_T s_T + T_c b_2 c_T s_T - b_1 b_2 s_T^2} \quad (21)$$

Equation (21) is the solution of the period-2 gait. And the step lengths of the first and the second step are

$$d_1 = x_1^- - x_1, d_2 = x_2^- - x_2 \quad (22)$$

where

$$x_1^- = x_0 c_T + T_c v_0 s_T$$
$$x_2^- = x_1 c_T + T_c v_1 s_T \quad (23)$$

Substituting (23) into (22) gives

$$d_1 = \frac{-2T_c a_1 s_T^2 + 2a_1 b_2 c_T s_T - 2a_2 b_1 s_T}{-T_c s_T^2 + b_1 c_T s_T + b_2 c_T s_T - b_1 b_2 s_T^2 / T_c}$$
$$d_2 = \frac{-2T_c a_2 s_T^2 + 2a_2 b_1 c_T s_T - 2a_1 b_2 s_T}{-T_c s_T^2 + b_1 c_T s_T + b_2 c_T s_T - b_1 b_2 s_T^2 / T_c} \quad (24)$$

When using different parameters, the values of $d_1, d_2$ have many possibilities, which represents various walking styles. For example, when $d_1, d_2$ are both positive, the robot goes forward; when $d_1, d_2$ are both negative, the robot goes backward; when they are one positive and one negative, the robot will take one step forward and one step backward. Particularly, there are two special cases of the period-2 gait:

(1) $a_1 = a_2 = a$, $b_1 = b_2 = b$

In this case, the period-2 gait turns into the period-1 gait. It can be easily verified that $d_1, d_2$ in (24) both become equal to $d$ defined in (17).

(2) $a_1 = a$, $a_2 = -a$, $b_1 = b_2 = b$

In this case, we have

$$d_1 = \frac{2T_c a(c_T + 1)}{T_c + T_c c_T - b s_T}, \quad d_2 = -d_1 \quad (25)$$

Since $d_1 + d_2 = 0$, it indicates that the solution is an in-place walking gait. With a bigger $a$, the two legs will be apart farther.

### III. LINEAR FOOT PLACEMENT CONTROL IN 3D

The LIP walking model in 3D is as follows [11]

$$\ddot{x} = gx/h$$
$$\ddot{y} = gy/h \quad (26)$$

This model is a combination of the 2D models in x and y directions, respectively. And the two directions are decoupled, thus they can be controlled independently.

If we select the following foot placement controller for x-axis

$$x_{f_1} = a_w + bv_x$$
$$x_{f_2} = -a_w + bv_x \quad (27)$$

and the following controller for y-axis

$$y_{f_1} = a_l + bv_y$$
$$y_{f_2} = a_l + bv_y \quad (28)$$

where $x_{f_1}, x_{f_2}$ represent the foot location in x-axis for leg 1 and leg 2, respectively; $y_{f_1}, y_{f_2}$ represent the foot location in y-axis for leg 1 and leg 2, respectively; $a_w, a_l$ and $b$ are controller parameters. It results in the 3D normal walking gait in y-direction as shown in Fig. 4, which can be characterized by the step length $d_l$ and step width $d_w$. Moreover, the step length and step width are proportional to $a_l$ and $a_w$, respectively, as can be verified from (17) and (25).

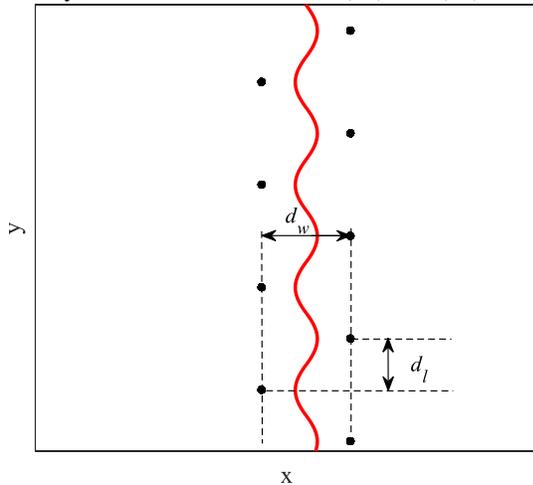

Figure 4. The 3D normal walking gait in y direction. The black dots are the footprints, and the red line is the CoM trajectory. The gait can be described by the step length $d_l$ and the step width $d_w$.

When considering the walking direction, one more gait parameter needs to be introduced. Denote the angle between the walking direction and the +y direction as $\theta$. Then 3D normal walking can be described by a triple $(d_l, d_w, \theta)$. To generate this gait, we need to modify the controller. This can be achieved by using the rotation matrix, and the resulted controller is as follows

$$x_{f_1} = -a_l \sin\theta - a_w \cos\theta + bv_x$$
$$x_{f_2} = -a_l \sin\theta + a_w \cos\theta + bv_x$$
$$y_{f_1} = -a_l \cos\theta + a_w \sin\theta + bv_y \quad (29)$$
$$y_{f_2} = -a_l \cos\theta - a_w \sin\theta + bv_y$$

In this controller, we can select a desired step period $T$ and choose $b$ as the dead-beat parameter. Then it turns into a three-parameter controller with the triple $(a_l, a_w, \theta)$ to be selected, where each parameter determines the corresponding gait parameter in $(d_l, d_w, \theta)$, respectively. With the controller (29), we can adjust the step length, step width, and walking direction in 3D walking conveniently.

### IV. SIMULATION STUDY

Simulations are performed in Matlab with the model parameters $g = 10m/s^2, h = 1m$. The step period $T = 0.3$ is used. For this model, we can calculate that $T_c = 0.3162$, $b_{max} = 0.7159$, $b_{db} = 0.4278$, $b_{cp} = 0.3162$, $b_{min} = 0.1397$.

#### A. 2D Walking Simulation

(1) Impact of parameter $b$

Here we use different values of $b$ in the controller. The initial state is set to $x_0 = -0.3, v_0 = 2$, and the parameter $a$ is set to zero. The resulted speed curves are shown in Fig. 5.

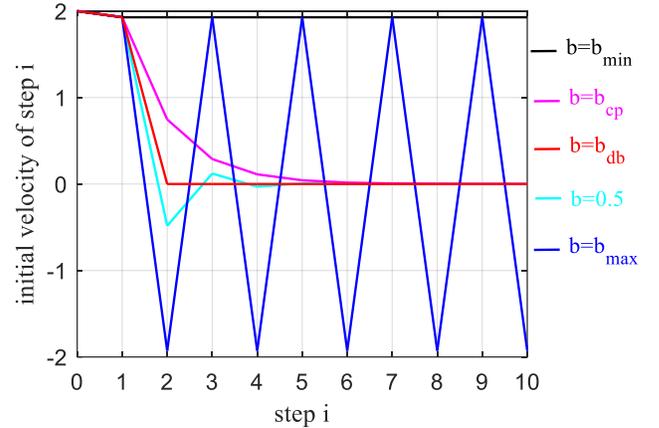

Figure 5. Speed curves for different values of b.

It should be noted that the foot placement control happens in the end of each step. So the first step is uncontrolled and the velocity changes the same in the beginning. After the first step, we have the following observations: 1) when $b = b_{min}$, the velocity maintains the same in each step; 2) when $b = b_{cp}$, the velocity converges to zero gradually without oscillation; 3) when $b = b_{db}$, the velocity converges to zero in one step; 4) when $b = 0.5$, the velocity converges to zero gradually with oscillation; 5) when $b = b_{max}$, the velocity has a continuous oscillation. The results are consistent with the previous analysis.

(2) Impact of parameter $a$

Here we fix $b$ to 0.3 and the initial condition $x_0 = 0, v_0 = 0$. Four cases are studied using different values of $a$ ($a_1$ for leg 1 and $a_2$ for leg 2). We calculate for 20 steps of walking and show the stance leg trajectory in the last 4 steps in Figs. 6-9. Case 1 leads to normal walking; Case 2 leads to in-place walking; Case 3 leads to asymmetrical walking where leg 2 takes smaller step than leg 1; Case 4 leads to asymmetrical walking where leg 2 takes smaller step in the opposite direction of leg 1. It demonstrates that a rich walking style can be generated by using different values of $a$.

- **Case 1**: Normal walking, $a_1 = a_2 = 0.2$ (periodic solution: $d_1 = d_2 = -0.35$)

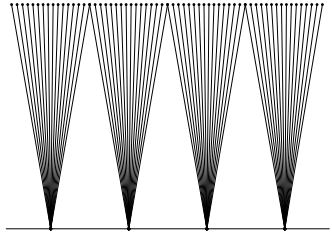

Figure 6. Stance leg trajectory of the last 4 steps in Case 1.

- **Case 2**: In-place walking, $a_1 = 0.2, a_2 = -0.2$ (periodic solution: $d_1 = 0.69, d_2 = -0.69$)

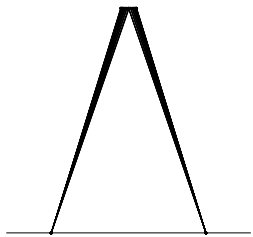

Figure 7. Stance leg trajectory of the last 4 steps in Case 2.

- **Case 3**: Asymmetrical walking, $a_1 = 0.2, a_2 = 0.4$ (periodic solution: $d_1 = -0.87, d_2 = -0.18$)

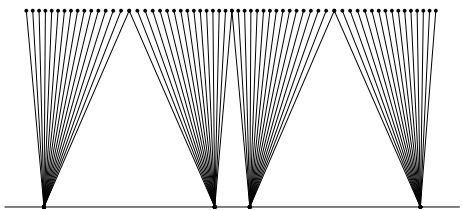

Figure 8. Stance leg trajectory of the last 4 steps in Case 3.

- **Case 4**: Asymmetrical walking, $a_1 = 0.2, a_2 = -0.4$, (periodic solution: $d_1 = 1.2, d_2 = -0.86$)

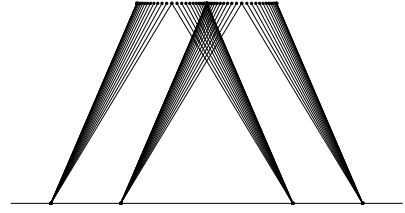

Figure 9. Stance leg trajectory of the last 4 steps in Case 4.

*B. 3D Walking Simulation*

For 3D walking, we first investigate straight walking with different values of $a_l, a_w$, which results in walking gaits with different step length and step width as shown in Fig. 10. Then we study walking with direction changes. Two cases are shown in Fig. 11 and Fig. 12, respectively. As can be seen from Fig. 11, the foot will be away from the CoM when turning happens. And for a sharper turn, the foot needs to step farther away. Since the leg length is limited for a practical robot, sharp turning should be avoided. Instead, the robot can take multiple mild turns continually to achieve a big turn, just like is shown in Fig. 12. The walking animation of Fig. 11 and Fig. 12 can be found in the video attachment.

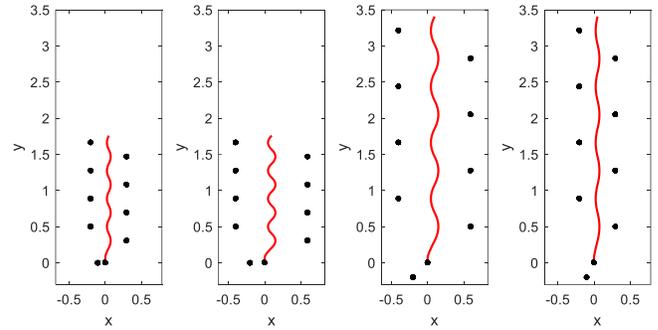

Figure 10. 3D gaits for different parameters. Left to right: (1) $a_l = 0.2, a_w = 0.1$; (2) $a_l = 0.2, a_w = 0.2$; (3) $a_l = 0.4, a_w = 0.2$; (4) $a_l = 0.4, a_w = 0.1$. It can be seen that (1) and (2), (3) and (4) have the same step length, while (1) and (4), (2) and (3) have the same step width.

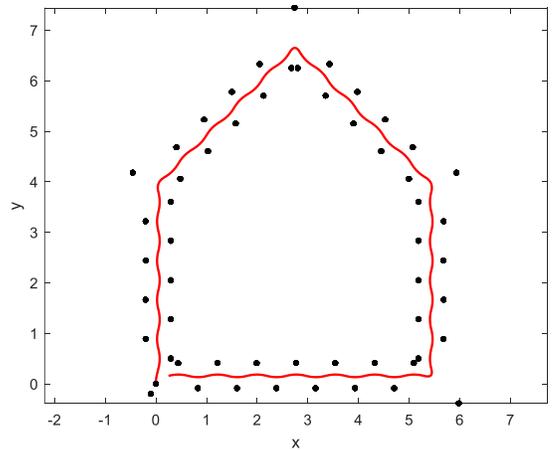

Figure 11. 3D walking with turning. The walker starts walking along +y from the origin with zero speed and turns four times of 45°, 90°, 45°, and 90°, respectively.

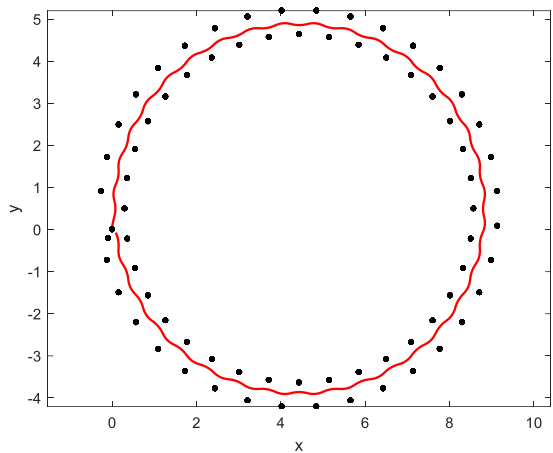

Figure 12. 3D walking in a circle. The walker starts walking along +y from the origin with zero speed and turns for 10° after each two steps.

*C. Discussions*

Though the previous theoretical analysis and simulations have shown the feasibility of LFPC on an ideal LIP model, it is still of great concern about how to implement LFPC on a real biped robot. While LFPC has specified the foot placement control of the swing leg, a more complete walking controller should also include the stance leg control which involves complicated joint torque calculations to maintain foot contact, body height, and body attitude. A successful walking can only be achieved as all of the control loops work well together.

As a simple verification, we build simple biped models in the V-REP [19] software as shown in Fig. 13. The models have a cuboid body, spherical hip joints, and telescopic legs ended with small feet. Despite the LFPC applied to the swing leg, controllers for the stance leg are also designed, including the linear joint controller to maintain a constant body height and the hip joint controller to keep the body upright. Those controllers are nearly decoupled, just like the three-part controller designed by Marc Raibert [2].

We investigate both 2D (by constraining the robot between two frictionless walls) and 3D walking and test the controller in various situations, including using different values for the parameter $a, b$, adding external forces, and walking on stairs and slopes. The simulation results are given in the video attachment. Overall, the simulation results have shown high consistency with the previous theoretical analysis, which verifies the effectiveness of the LFPC strategy in more realistic situations.

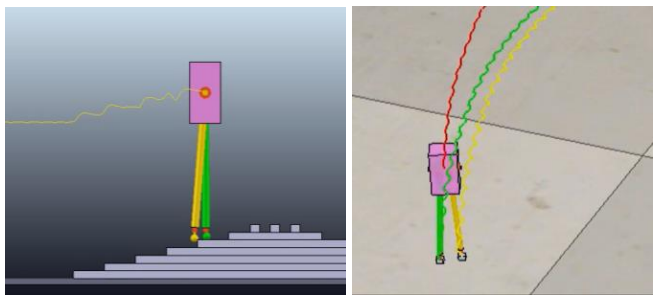

Figure 13. V-REP simulation scenes. Left: 2D walking. Right: 3D walking.

## V. CONCLUSIONS

Foot placement is the most important way to maintain balance for humans and biped robots. Despite some scattered results, the question of "when and where to take a step" has not yet be fully addressed. In this paper, we investigate the simplest balance controller for dynamic walking which uses a linear foot placement with respect to velocity. LFPC has three parameters $T, a, b$, where $T$ tells when to take a step and $a, b$ specifies a linear function of the body velocity that determines where to take a step. Theoretical analysis of LFPC on the LIP model gives the balance criterion and also discovers the property of "faster stepping, easier to balance". Particularly, there exists a special value of $b$ that leads to an immediate convergence to a steady-state gait in just one step. LFPC can also be easily extended to 3D and we have proposed a simple three-parameter controller that can control the step length, step width as well as walking direction intuitively.

The results in this work can be applied to both biped and quadruped robots. However, some practical factors such as leg length constraint, joint velocity constraint, and leg mass are not considered here for the sake of not adding complexity. It is necessary to do further study on their impacts in the future. Besides, we are developing a biped robot "Tik-Tok" [20] together with Professor Andy Ruina and we hope the proposed controller can be applied to our robot when it is completed.